\documentclass[sigconf,10pt]{acmart}
\renewcommand\footnotetextcopyrightpermission[1]{}
\renewcommand\footnotetextcopyrightpermission[1]{} %
\setcopyright{none}
\usepackage{booktabs}
\usepackage{multirow}
\usepackage{float}    %
\usepackage{graphicx} %
\usepackage{subcaption} %
\usepackage{xspace}   %
\usepackage{makecell} %
\usepackage{tikz}     %
\usetikzlibrary{positioning,fit,backgrounds,decorations.pathreplacing,calc,arrows.meta,matrix,shapes.geometric}
\usepackage{algorithm} %
\usepackage{algpseudocode} %

\def\myname{SchemaWeaver\xspace}
\begin{document}

\title[\myname{}: Accelerating On-Device Mobile Agentic AI via Schema-Guided Parallel LLM Decoding]{\myname{}: Accelerating On-Device Mobile Agentic AI via Schema-Guided Parallel LLM Decoding}

\author{Boyuan Yang}
\affiliation{%
	\institution{University of Pittsburgh}
	\city{Pittsburgh}
	\state{PA}
	\country{USA}
}
\email{by.yang@pitt.edu}

\author{Wei Gao}
\affiliation{%
	\institution{University of Pittsburgh}
	\city{Pittsburgh}
	\state{PA}
	\country{USA}
}
\email{weigao@pitt.edu}

\begin{abstract}
Large language models (LLMs) on mobile devices can act as agents that call external software tools to complete complex tasks, but the major challenge of supporting on-device mobile agentic AI with LLMs is the long latency of LLM tool calling on mobile devices. In particular, the decode phase is the main contributor of such long latency, but is hard to speed up on mobile processors as it is bound by memory bandwidth rather than compute power. To address this limitation, in this paper we present \myname, a new on-device mobile agentic AI technique that enforces schema-guided parallel decoding in the decode phase of LLM tool calling, to reduce its latency by producing multiple tokens in each decode pass. \myname can be generically applied to the LLM inference engine on mobile devices without involving any training efforts or changes on LLMs, and our experiment results on major mobile SoCs and LLM families show that \myname can reduce the LLM tool calling latency by up to 3.9 times, compared to the competitive baselines.
\end{abstract}

\maketitle

\vspace{-0.05in}
\section{Introduction}
\label{section:introduction}
\begin{figure}[t]
	\centering
	\includegraphics[width=0.95\columnwidth]{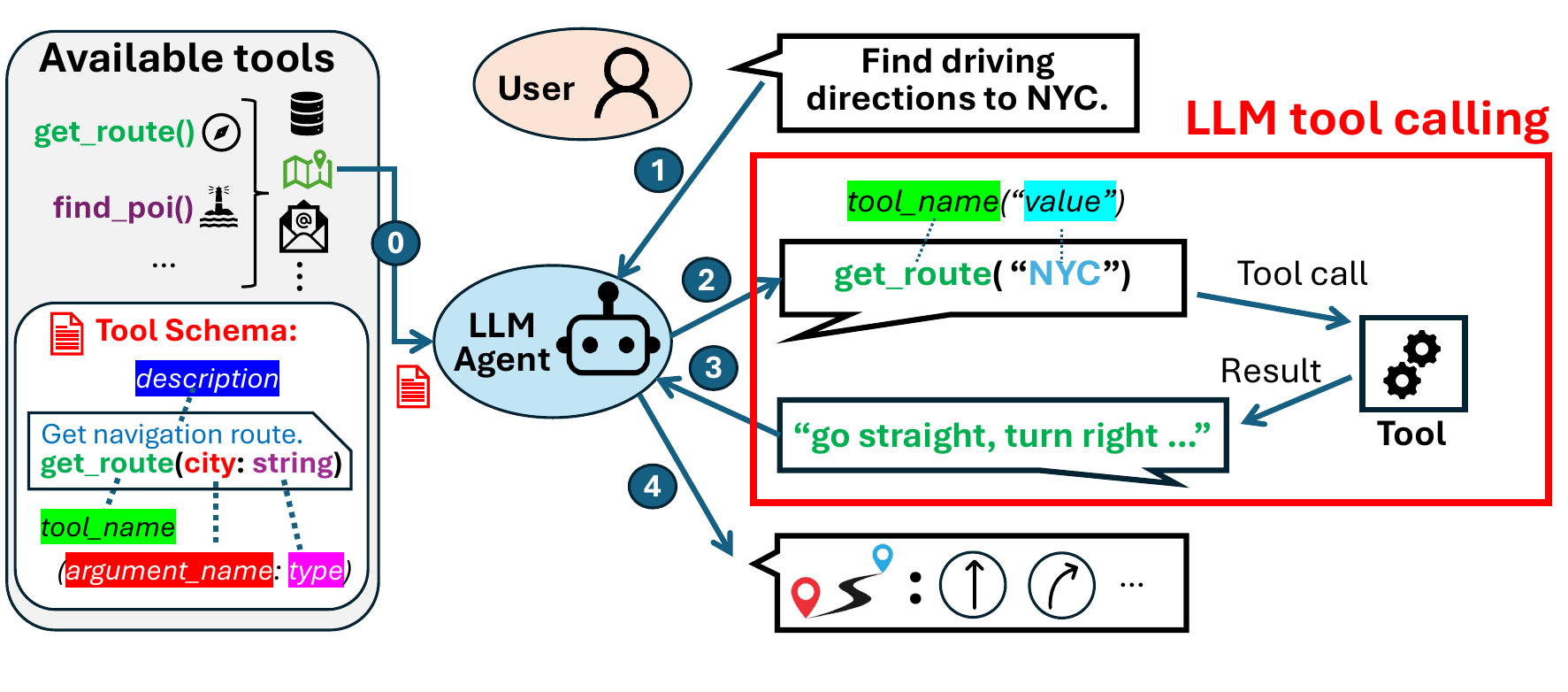}
	\vspace{-0.1in}
	\caption{Agentic AI systems based on LLM agents that call external software tools}
	\vspace{-0.2in}
	\label{fig:toolcalling}
\end{figure}

In recent years, Large Language Models (LLMs) have been used as the backbone of \emph{agentic AI} systems. When given proper context, LLM agents can interpret user instructions, select the proper external software tools, and call these tools to complete complex tasks \cite{schick2023toolformer, yao2023react, qin2024toolllm, hou2025model}. For example in Figure \ref{fig:toolcalling}, when being asked for driving directions, the LLM agent selects the \texttt{get\_route} software tool from the available tools for this task. It then calls this tool for route planning and returns the results back to the user in natural language. In this process, the LLM agent perceives each tool from its \emph{schema}, which is a machine-readable description that gives the tool's description and the names and types of its arguments \cite{hou2025model}. The LLM agent, then, calls the tool by generating the text command with values of the tool's arguments.

Currently, LLM agents on mobile devices (e.g., smartphones) have become increasingly popular, %
allowing intelligent and automated operations on the user's personal data \cite{chen2025llm,nong2024mobileflow,wen2024autodroid}. Agentic AI functionalities on mobile devices have also been supported by mainstream mobile OSes, via Android AppFunctions~\cite{android2026appfunctions, android2026intelligentos} and iOS App Intents~\cite{apple2024appintents}. In particular, running on-device LLM agents on local devices well preserves the user's data privacy \cite{xu2024ondevicereview, yin2024llmaas, li2024personal}, and quantized LLMs with 1B-3B parameters, which fit in a smartphone's memory budget \cite{liu2024mobilellm, frantar2023gptq, lin2024awq, llamacpp}, have been capable of launching tool calls with reasonable accuracy \cite{liu2024xlam, paramanayakam2024lessismore, haque2025tinyllm}.

\begin{figure*}[t]
	\centering
	\vspace{-0.1in}
	\hspace{-0.1in}
	\begin{subfigure}[b]{0.54\textwidth}
		\centering
		\includegraphics[width=\linewidth]{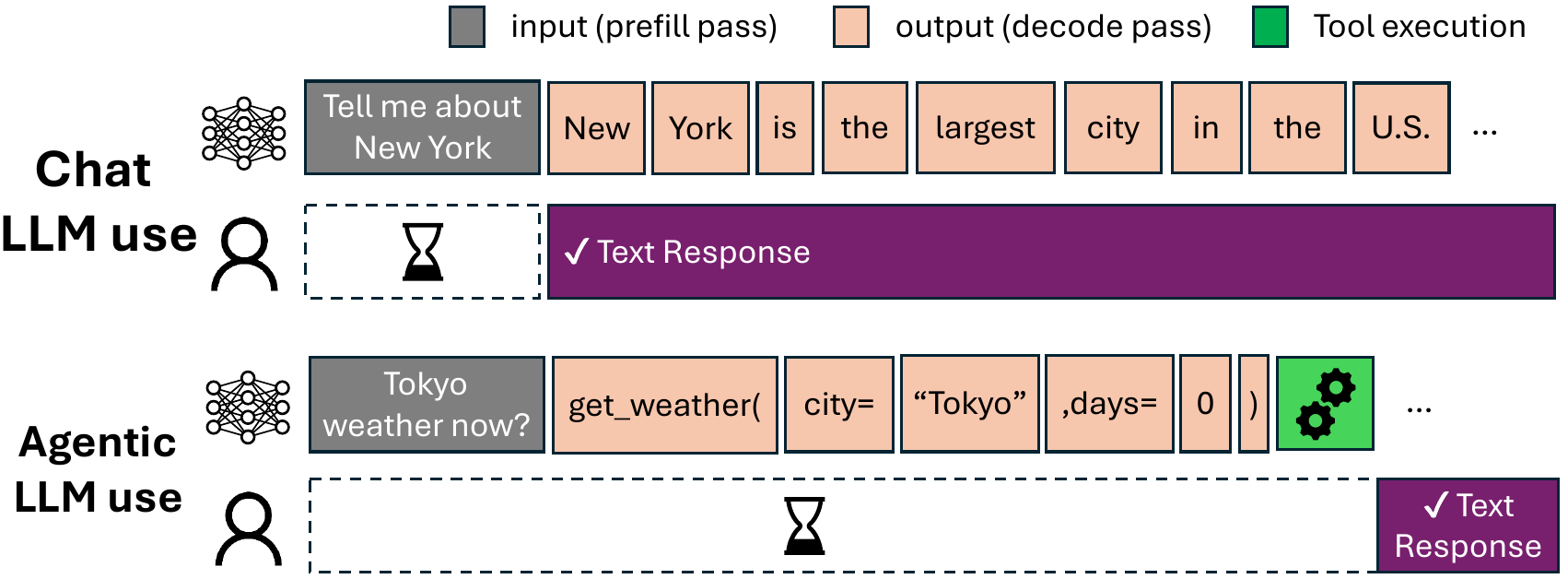}
		\caption{Agentic LLM use incurs longer latency to the user}
		\label{fig:overview-a}
	\end{subfigure}
	\hfill
	\begin{subfigure}[b]{0.46\textwidth}
		\centering
		\includegraphics[width=\linewidth]{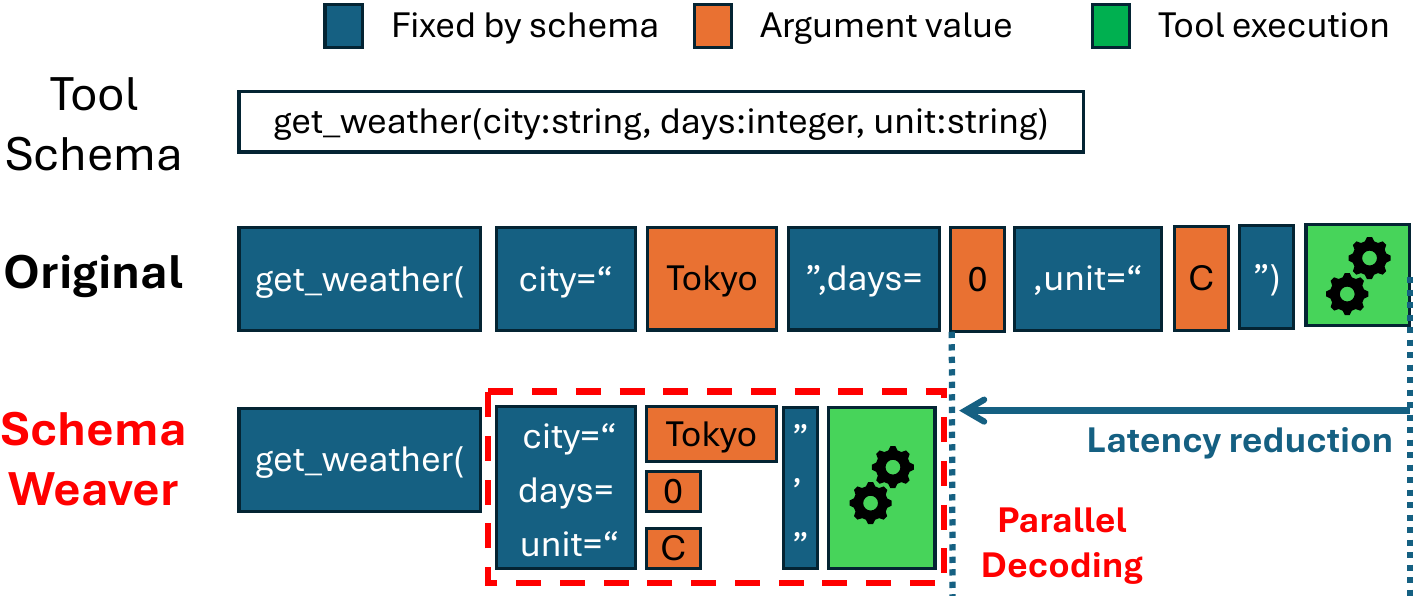}
		\caption{\myname{} reduces latency via parallel decoding}
		\label{fig:overview-b}
	\end{subfigure}
	\hspace{-0.1in}
	\vspace{-0.15in}
	\caption{The latency of LLM tool calling in on-device mobile agentic AI}
	\vspace{-0.15in}
	\label{fig:overview}
\end{figure*}

The major challenge of supporting on-device mobile agentic AI, however, is the long latency of LLM tool calling on mobile devices. %
In inference, an LLM produces its output through repeated forward passes that constitute a \emph{prefill phase} and a \emph{decode phase}. The prefill phase uses the first pass to read the input prompt, and the decode phase then generates the response using the other passes, each of which produces one output token \cite{pope2023scaling, xu2025mllmnpu}. As shown in Figure~\ref{fig:overview}(a), in traditional conversation-based LLM uses, one prefill is usually followed by a large quantity of decoded text tokens that the user reads as they arrive, and most of the decode latency is hence ``hidden'' in user reading \cite{liu2024andes, xiao2025cognitiveload}. Agentic LLM operations, in contrast, first decode a structured tool call command for tool execution, and text tokens in response can only be produced after the completion of tool execution. The decode latency experienced by the user, hence, becomes significantly longer, especially when executing complicated agentic tasks such as on-device content search that involves multiple tool calls of different types. In these tasks, our experiments show that the decode latency constitutes up to 81\% of the wall-clock time being taken (Sec \ref{subsec:latency_breakdown}).

Unfortunately, it is difficult to accelerate this decode phase on mobile devices, which is bound by the device's limited memory bandwidth rather than by compute power \cite{pope2023scaling, huang2025mnnaecs}. More specifically, on mobile devices, the LLM's weights have to be repeatedly loaded from memory to mobile processors by every pass, due to the processors' limited cache capacities \cite{pope2023scaling, alizadeh2024llmflash}. Since each decode pass only produces one output token, the decode latency is determined by the speed of memory bus, which is usually much slower than the mobile processor speed and hence causes processor starvation.
Our experiment results on the OnePlus 12R phone with Snapdragon 8 Gen 2 SoC show that the decode speed on mobile CPU is $>5\times$ slower than the prefill speed (Sec \ref{subsec:bandwidth}). Using mobile GPU or NPU cannot help much, due to the shared memory architecture in mobile SoCs \cite{chen2025llm, xu2025mllmnpu, li2026npunotfaster, cai2026npuready}.

The key to accelerating the decode phase in mobile agentic AI, hence, is how to allow producing multiple output tokens in one decode pass, to fully utilize the on-device compute power. One straightforward approach is speculative decoding~\cite{leviathan2023speculative, cai2024medusa, fu2024lookahead}, which first uses a much smaller draft model to guess the next few tokens and then lets the target model check the correctness of these guesses in one pass. Since a tool call is highly structured, most output tokens are highly predictable. For example, with a SmolLM2-1.7B as the target model and SmolLM2-135M as the draft model, 74\% of guesses are correct. However, these correct guesses only result in 31\% speedup, because the draft model still decodes one token per pass and is bounded by memory bandwidth as well.

Another option is constrained decoding~\cite{xgrammar2, dong2024xgrammar, willard2023outlines, geng2023gcd, zheng2024sglang}, which builds a set of grammatical rules from the tool's schema and forbids any output token that makes the LLM tool call ill-formed. These rules can be further used by jump-forward decoding \cite{zheng2024sglang} in each decode pass, to skip the current decode computation when only one legitimate output token exists. These skipped tokens can then be processed in batch by a single prefill pass. However, since tokens can only be skipped when they are being computed, the overall decode latency is still bound by the number of decode passes, which can only be sequentially computed.

To fundamentally overcome this limitation of sequential decode computations, we instead aim to allow decoding the argument values of an LLM tool call in parallel, by leveraging the pre-known and fixed information contained in the tool schema. More specifically, the schema describes the position of each argument's value in the tool call command, and each position is always fixed. For example, in the call \texttt{get\_weather()} shown in Figure~\ref{fig:overview}(b), the value of argument \texttt{city} is a string that begins after \texttt{city="} and ends at the closing quote, and the value of argument \texttt{days} is an integer after \texttt{days=}. Hence, instead of decoding these argument values sequentially, we can decode all of them as one batch in parallel to save unnecessary passes.
On the other hand, the fixed names of arguments need no decoding, and can instead be loaded by the LLM inference engine via one prefill pass.

Based on this idea, in this paper we present \myname{}, a new on-device mobile agentic AI technique that enforces schema-guided parallel decoding in the LLM inference engine. First, \myname parses the fixed parts in the tool schema to build a \emph{tool skeleton} that provides information about the position of each argument value (Sec \ref{subsection:call-skeleton}). It then places argument names as parallel token sequences, so that arguments share the same prompt and tool call prefix to facilitate parallel decoding of argument values. To handle rugged value lengths, we pad decode sequences on the fly for the maximum compute efficiency (Sec \ref{subsection:value-streams}). Since decode results of argument values may not share exactly the same sequential prefix, we add a write-back phase after parallel decoding to realign the decode results and rebuild the LLM agent's KV cache accordingly. To avoid this write-back phase from blocking tool executions, we assemble a valid tool call at the same time and emit the tool call early to further reduce the end-to-end latency (Sec \ref{subsection:call-assembly}).

Many tools called by the LLM agent may include optional arguments. For example, when calling the weather query tool, the LLM agent may opt to specify the temperature unit or not. The values of optional arguments have to be decided after all required argument values are generated, and cannot adapt to parallel decoding.
Instead, we have to decide and generate values of optional arguments in a sequential manner, after parallel decoding of required argument values (Sec \ref{subsection:late-decision}). To minimize the extra latency caused by this sequential decoding, we design provisional calls for read-only tools to emit early only with required arguments, which can be later replaced by a corrected call once the LLM agent adds the optional argument values (Sec \ref{subsection:provisional-calls}).

To our best knowledge, \myname{} is the first system design that accelerates on-device LLM tool calling on mobile devices, without involving any training efforts or changes on LLMs. We implemented \myname{} in llama.cpp~\cite{llamacpp} inference engine by using multiple tokens as a matrix in decoding when performing GEMM (general matrix-matrix multiply)\footnote{Our implementation achieves parallel decoding by increasing the data parallelism, and can hence be generically applied to different mobile processors including CPU, GPU and NPU.}, and evaluated it on multiple tool calling benchmarks, mobile SoCs and LLM families including Qwen~\cite{yang2024qwen25, yang2025qwen3}, Llama~3.2~\cite{grattafiori2024llama3} and SmolLM2~\cite{allal2025smollm2}. Based on experiment results, we have the following conclusions:

\vspace{-0.05in}
\begin{itemize}
	\item \myname{} reduces the number of forward passes for a tool call from 44.3 to 13.1, and the end-to-end tool calling latency by $2.13\times$ over the competitive baselines.
	\item The gain of \myname{} grows as the memory bandwidth falls. When the memory bus is slowed by $4.7\times$, serial decoding slows by $3.2\times$ but \myname{} only slows by $2.4\times$, and its speedup rises from $3.0\times$ to $3.9\times$.
	\item Applying \myname{} costs $<$2\% of tool calling correctness across different LLM settings.
	\item \myname{} is lightweight and efficient. With 8.9\,MB of extra peak memory being used compared to baselines, it reduces the energy consumption of a tool call by $2.1\times$.
\end{itemize}

\vspace{-0.1in}
\section{Background \& Motivation}
\label{section:background}

\subsection{Tool Calling Interfaces of LLM Agents}
\label{section:tool-call-interface}

Nowadays, different tool calling interfaces are available and they differ in how a tool is presented to and executed by the LLM agent. Model Context Protocol (MCP) \cite{hou2025model, mcp2025spec} is the most commonly used and provides tool schemas in JSON format with key-value pairs, such as \texttt{\{"name": "get\_weather", "arguments": \{"city": "Tokyo"\}\}}. Other formats also exist, such as the Python-like tool calls in Llama 3.2 \cite{grattafiori2024llama3} and the special tokens around each argument value in FunctionGemma~\cite{google2025functiongemma}. In each format, the tool schema fixes every character of the tool call except the argument values, and \myname hence generally applies to all formats.

\begin{figure}[ht]
	\centering
	\vspace{-0.1in}
	\includegraphics[width=0.85\columnwidth]{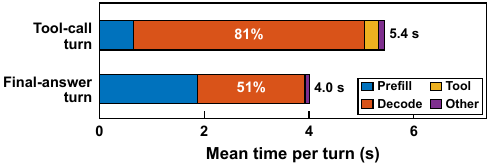}
	\vspace{-0.15in}
	\caption{Wall-clock time breakdown of LLM tool calling latency on an OnePlus 12R smartphone}
	\vspace{-0.1in}
	\label{fig:breakdown}
\end{figure}

Some desktop coding agents allow the LLM agent to act in free text instead of following the tool schema, by generating a shell command or a program~\cite{yang2024sweagent, wang2024codeact}. Such an action is passed through a tool with a single text argument, e.g., \texttt{\{"command": "ls -l"\}}, or executed directly as code. %
These agents, however, cannot apply to mobile agentic AI, because a mobile OS often exposes the app functionality through typed and permission-controlled interfaces such as Android AppFunctions~\cite{android2026appfunctions, android2026intelligentos} and iOS App Intents~\cite{apple2024appintents}. Hence, in the rest of this paper, we consider an LLM tool call to always include a tool name with typed argument names and values.

\vspace{-0.05in}
\subsection{Tool Call Latency Breakdown}
\label{subsec:latency_breakdown}

The breakdown of LLM tool calling latency, on a OnePlus 12R smartphone with Qwen3-1.7B-Q8 model, is shown in Figure \ref{fig:breakdown}, where the tool-call turn emits the tool call command from the user input (e.g.,  \textcircled{1} and \textcircled{2} in Figure~\ref{fig:toolcalling}) and the final-answer turn executes the tool and generates response to the user based on results of tool execution (e.g.,  \textcircled{3} and \textcircled{4} in Figure~\ref{fig:toolcalling}). In both cases, decode contributes the most (up to 81\%) to the end-to-end tool calling latency. In particular, a large amount of decode computation in the tool-call turn is redundant: our analysis of 2,351 tool calls in the Berkeley Function Calling Leaderboard (BFCL-v4)~\cite{patil2025bfcl} dataset shows that, for any selected tool, 54-63\% of a tool call's tokens in the decode phase of the tool-call turn are fixed. Hence, \myname{} mainly targets the tool-call turn and reduces its latency by eliminating redundant decode computations.

\begin{table}[ht]
	\centering
	\small
	\vspace{-0.05in}
	\begin{tabular}{llrr}
		\toprule
		Mobile Processor & Prefill & Decode \\
		\midrule
		Hexagon NPU & \textbf{685.4} & 16.4 \\
		Adreno GPU & 93.1 & 11.3 \\
		CPU & 145.7 & \textbf{22.6} \\
		\bottomrule
	\end{tabular}
	\vspace{0.05in}
	\caption{Prefill and decode throughputs (tokens/s) on mobile processors with Qwen3-1.7B-Q8 model.}
	\vspace{-0.2in}
	\label{tab:backends}
\end{table}

\begin{figure}[ht]
	\centering
	\vspace{-0.1in}
	\begin{subfigure}[t]{0.5\columnwidth}
		\centering
		\includegraphics[width=\linewidth]{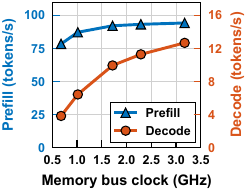}
		\caption{Throughput}
		\label{fig:batch-a}
	\end{subfigure}\hfill
	\begin{subfigure}[t]{0.45\columnwidth}
		\centering
		\includegraphics[width=\linewidth]{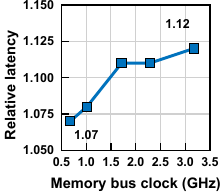}
		\caption{Relative decode latency}
		\label{fig:batch-b}
	\end{subfigure}
	\vspace{-0.1in}
	\caption{Prefill and decode performance with different memory bus frequencies on mobile devices}
	\vspace{-0.1in}
	\label{fig:batch}
\end{figure}

\begin{figure*}[t]
	\centering
	\includegraphics[width=\textwidth]{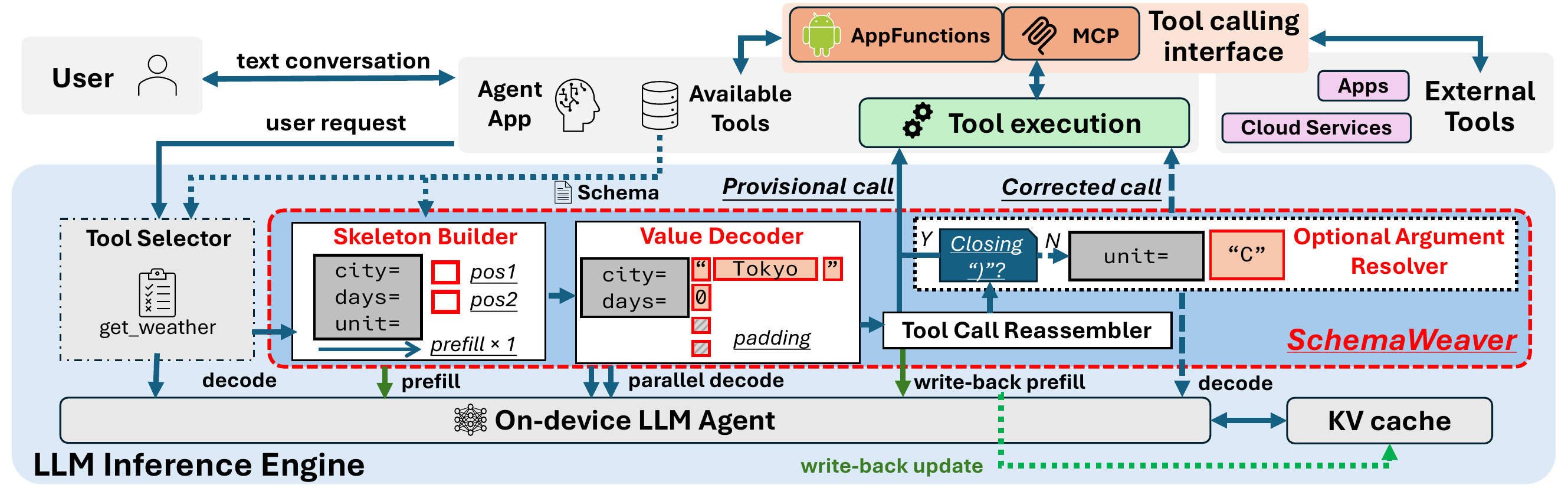}
	\vspace{-0.3in}
	\caption{Overview of \myname{} design}
	\vspace{-0.1in}
	\label{fig:design}
\end{figure*}

\vspace{-0.05in}
\subsection{Memory-Bound Tool Call Decoding on Mobile Processors}
\label{subsec:bandwidth}

Table~\ref{tab:backends} shows the prefill and decode throughput on a OnePlus 12R smartphone with Snapdragon 8 Gen 2 SoC. Among all on-device processors, CPU is the fastest for decoding, and the decode throughput is much slower than the prefill throughput on all mobile processors. These results match the findings in the existing work \cite{xu2025mllmnpu, chen2025llm}, and are caused by the mismatch between memory bus bandwidth and processor workloads. While prefill produces multiple tokens in one pass, one decode pass only produces one token and results in low utilization of mobile processors awaiting data from memory between passes. Our experiment results in Figure~\ref{fig:batch}(a) further support this observation: when we slow down the memory bus frequency, the decode throughput falls sharply by 70\% but prefill throughput falls by only 17\%.

Such a bound on memory bandwidth, however, creates unique opportunities for parallel decoding: if a decode pass already reads all the necessary data (LLM model weights, input prompt and tool schema) from memory, the extra latency incurred by decode computations of producing multiple tokens in this pass would be only marginal. As shown in Table~\ref{tab:batchcost}, when a decode pass outputs 4 tokens\footnote{The optimal sizes of token batch are multiples of 4, because LLM inference engine's CPU matrix routines handle 4 sequences at a time in a pass~\cite{llamacpp}.} instead of 1, the decode latency only increases by up to 20\% and the average per-token decode latency hence drops by 70\%. Figure \ref{fig:batch}(b) further shows that this saving is insensitive to changes of memory bandwidth.
These results motivate the parallel decoding design in \myname, to fully utilize the available compute power in decode passes by producing multiple tokens in each decode pass.

\begin{table}[ht]
	\centering
	\small
	\vspace{-0.05in}
	\begin{tabular}{@{}lrrrrrrr@{}}
		\toprule
		\# tokens in one pass & 1 & 2 & 3 & 4 & 5 & 6 & 8 \\
		\midrule
		OnePlus 12R & 1.00 & 1.61 & 2.15 & \textbf{1.20} & 1.91 & 2.49 & 2.26 \\
		Google Pixel 7 & 1.00 & 1.75 & 2.34 & \textbf{1.11} & 1.88 & 2.47 & 2.06 \\
		\bottomrule
	\end{tabular}
	\caption{Relative increase of the parallel decoding cost in one decode pass on different smartphone models}
	\vspace{-0.3in}
	\label{tab:batchcost}
\end{table}

\section{Overview}
\label{section:overview}
As shown in Figure \ref{fig:design}, \myname{} decodes the argument values of an on-device agentic AI tool call in parallel, and leaves everything fixed in the tool schema as prefill. To that end, we assume that every available tool comes with a schema that lists its arguments, types of arguments, and whether each argument is required or optional. This availability of tool schemas is guaranteed in most existing agentic frameworks, such as LangChain and AutoGen~\cite{langchain2022, wu2023autogen}.

\myname{} sits inside the on-device LLM inference engine and does not alter other components outside of tool calling. Without loss of generality, in \myname design we assume that the software tool being called is decided by a specialized Tool Selector given the user request, and \myname hence focuses on accelerating the corresponding call of the selected tool. In practice, such tool selection has been recognized as an easy NLP task, on which a small LLM (e.g., Octopus-v2-2B~\cite{chen2024octopus} or TinyAgent-1.1B~\cite{erdogan2024tinyagent}) can succeed.

\subsection{Tool Skeleton Builder}
\label{section:skeleton}

As stated in Section \ref{section:introduction}, an argument value can only be decoded in parallel with others if the inference engine knows its position, but traditional serial decoding learns the position of an argument value only when decoding of the previous value has ended. \myname, instead, makes the positions of all argument values pre-known before decoding, by building a \emph{tool skeleton} that contains all the fixed parts of tool call, namely the tool name and names of required arguments, from the tool schema. Argument values remain unfilled in the tool skeleton, as exemplified in Figure \ref{fig:design} for the tool \texttt{get\_weather} with two arguments \texttt{city=} and \texttt{days=}. Positions of argument values can then be derived by tokenizing the texts in tool skeleton with the argument names laid out in parallel and letting the LLM agent read these tokens in one prefill pass: the first token of each argument value can be located immediately as after the last prefilled token of the corresponding argument name.

Special handling, however, is needed at the boundary between the tool skeleton and an argument value, because the model may write the last character in the skeleton and the value's first character as one fused token, which distorts the parallel decoding of argument values. More details about the tool skeleton and such special handling are in Section \ref{subsection:call-skeleton}.

 \vspace{-0.05in}
\subsection{Parallel Value Decoder}
\label{section:parallel}

In parallel decoding of argument values, each argument value corresponds to one token sequence that starts from the token corresponding to the argument's name in the tool skeleton. All these token sequences can be computed as one batch, and a token sequence stops when the corresponding argument value is completely decoded, depending on the argument's type. For example, a string ends with the closing quote, and a number ends at the first token that cannot continue as a number. Decoded argument values hence differ in length, and the decoder applies padding to token sequences when necessary. More details are in Section \ref{subsection:value-streams}.

The decode results cannot be directly used as input to the LLM agent, because each sequence contains only part of the tool call command (i.e., the corresponding argument and its value). The LLM agent cannot continue its inference on any sequence, as it cannot read the full text of tool call command. To address this issue, \myname writes the decoded argument values back to a sequential space in device memory after the parallel decoding completes, to reassemble the complete call text and enter it into the LLM's KV cache as one new sequence. The write-back costs a single prefill pass instead of decoding the call again token by token, since the assembled call can be handed to the tool for execution before write-back, such reassembling does not incur extra latency on tool execution. More details about such write-back are in Section \ref{subsection:call-assembly}.

\vspace{-0.05in}
\subsection{Optional Argument Resolver}
\label{section:optional}

Optional arguments involved in the specific tool call need to be separately handled. Only after the write-back of required argument values, \myname decides the optional arguments one at a time with the Optional Argument Resolver, to ensure the correctness of generated tool call command. To minimize the latency of such serial decisions, \myname starts the tool early as a \emph{provisional call} that carries only the required arguments, while the Optional Argument Resolver decides the optional arguments. If the tool call indeed involves optional arguments, these are later added towards a corrected call that replaces the provisional call.

Early execution of the provisional call is feasible only when the tool is read-only and its execution does not apply any changes to the mobile system. For example, a provisional call to the weather lookup tool illustrated in Figure \ref{fig:overview} is feasible, but other tools such as sending a message cannot be executed by a provisional call. More details are in Section \ref{section:late}.

\vspace{-0.05in}
\section{Parallel Decoding of Argument Values}
\label{section:required}

In this section, we will use an end-to-end example of the \texttt{get\_weather} tool call with two required arguments, \texttt{city} and \texttt{days}, to present details about parallel decoding of the tool call's argument values.

\begin{figure}[t]
	\centering
	\includegraphics[width=\columnwidth]{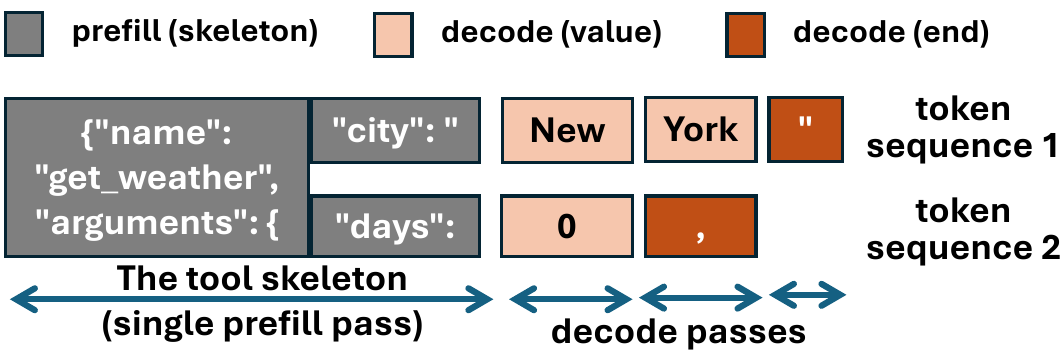}
	\vspace{-0.25in}
	\caption{The tool skeleton in one prefill pass, with every argument value starting at the same position}
	\vspace{-0.1in}
	\label{fig:skeleton}
\end{figure}

\vspace{-0.05in}
\subsection{Building the Tool Call Skeleton}
\label{sec:toolname}
\label{subsection:call-skeleton}

The tool skeleton is determined jointly by the tool schema and the LLM's tool call format being used, which fixes the tokens that open and close a call and the keys around the tool name and list of arguments. Different LLM families use different tool calling formats, and \myname's Tool Skeleton Builder adapts to them, by recognizing the format being used from the special tokens in the model's vocabulary. For example, for Qwen3 models \cite{yang2025qwen3}, the shared head of tool skeleton uses the opening \texttt{<tool\_call>} tag and the key \texttt{arguments}. The closing brace and the closing tag come after the last argument value, and so they are added at the write-back (Sec \ref{subsection:call-assembly}). For Llama 3.2 \cite{grattafiori2024llama3} models, the head has no tag and uses the key \texttt{parameters}.

Figure~\ref{fig:skeleton} shows how the tool skeleton is laid out. The shared head of the call lists the tool name and the beginning part of the argument list. They are shared and visible to all token sequences in decoding, each of which corresponds to one argument and starts at the same position after the shared head. A single prefill pass processes the head and all argument names together; afterwards, the starting position of every argument value (in decoding) is known, as decoding follows from the skeleton's next token in the sequence.

\begin{table}[ht]
	\centering
	\footnotesize
	\setlength{\tabcolsep}{3pt}
	\vspace{-0.1in}
	\begin{tabular}{l|c|c|c}
		\toprule
		\textbf{Model} & \makecell{\textbf{Serial decode}\\\textbf{(ms)}} & \makecell{\textbf{Jump-forward}\\\textbf{prefills (ms)}} & \makecell{\textbf{\myname's }\\\textbf{prefill (ms)}} \\
		\toprule
		Qwen3-1.7B & 1,201 & 577 & 314 \\
		Qwen3-1.7B 4-bit & 999 & 615 & 335 \\
		Qwen3-0.6B & 545 & 299 & 172 \\
		\midrule
		Llama-3.2-1B & 743 & 407 & 222 \\
		Llama-3.2-3B & 1,886 & 997 & 551 \\
		\midrule
		SmolLM2-1.7B & 1,352 & 945 & 395 \\
		\bottomrule
	\end{tabular}
	\vspace{-0.05in}
	\caption{Compute cost of tool skeleton decode per call}
	\vspace{-0.25in}
	\label{tab:skeleton}
\end{table}

Implementing the skeleton decode as prefill is beneficial. Table~\ref{tab:skeleton} shows the compute cost of building the skeleton on 100 tool calls picked from the BFCL dataset \cite{patil2025bfcl} using a OnePlus 12R smartphone, where \myname achieves up to $3.8\times$ speedup compared to other approaches.

In addition, %
most tool call formats insert a space in front of argument values, and the LLM tokenizers \cite{sennrich2016bpe} fuse the prefix space and the first character of the argument value, such as the minus sign (\texttt{-}), into the same token. Placing the space into the tool skeleton hence confuses the model into emitting wrong values. \myname{} builds the skeleton with the LLM tokenizer as the reference, to avoid splitting a fused token between the skeleton and argument values.

\begin{figure}[ht]
	\centering
	\vspace{-0.05in}
	\includegraphics[width=0.85\columnwidth]{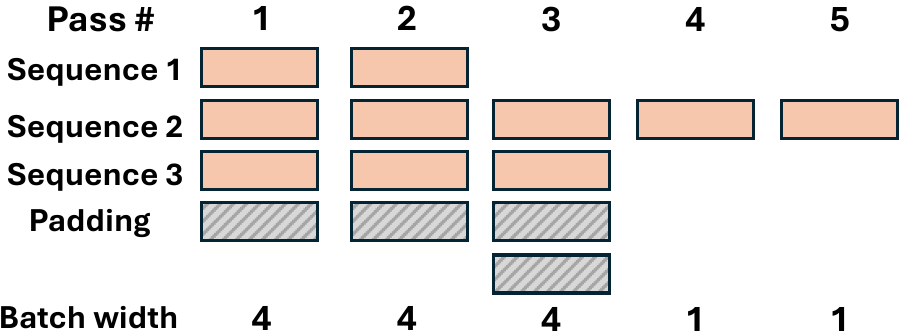}
	\vspace{-0.1in}
	\caption{Parallel decoding of three argument values with rugged lengths being padded}
	\label{fig:values}
\end{figure}

\vspace{-0.05in}
\subsection{Parallel Token Sequences}
\label{sec:streams}
\label{subsection:value-streams}

As shown in Figure \ref{fig:skeleton}, each token sequence is separated from others, and restricted to the type of its argument and ends where its type ends: at the closing quote of a string, at the separator after a number or a boolean, or at the matching bracket of an array or an object.

\noindent\textbf{Decoding pass reduction.} As shown in Figure \ref{fig:values}, the decoding for all token sequences advances together. For a call with $k$ required argument values $v_1, \dots, v_k$ of $|v_i|$ tokens each, serial decoding spends $\sum_i |v_i|$ decode passes, whereas \myname{} takes $s_{\text{name}} + \max\nolimits_{i} |v_i|$ decode passes plus two prefill passes, where $s_{\text{name}}$ corresponds to the one or two passes that decide the tool name, and the two prefill passes are for the tool skeleton and the write-back (Sec \ref{subsection:call-assembly}). When being tested on the BFCL dataset \cite{patil2025bfcl}, \myname{} reduces the number of decode passes from 13.1 to 6.8 on average, and reduces the decoding latency by 39\% accordingly. Especially for complicated tool calls with 2 or more required arguments, it takes advantage of argument parallelism and further reaches 47\% latency reduction.

\noindent\textbf{Padding rugged value lengths.} A pass with $n$ unfinished sequences computes $n$ tokens in one batch. However, decoding with the minimum number of sequences (i.e., only the unfinished sequences) is not always proper on mobile devices. As shown in Table \ref{tab:batchcost}, a batch of 4 minimizes the per-sequence decode cost as a result of mobile LLM inference engine (e.g., llama.cpp), and the Parallel Value Decoder in \myname hence pads a pass with two or three unfinished sequences using placeholder tokens. Our statistics show that, over tool calls in the BFCL dataset \cite{patil2025bfcl}, such padding is applied in 27\% of decode passes and further improves the decode speedup of \myname{} by another 17\%.

\noindent\textbf{Limited impact on correctness.} Parallel decoding may impact the LLM agent's correctness of generating the tool call command, because of the mutual separation between parallel token sequences. For example in a flight search task from Berlin to New York, serial decoding writes the origin as the airport code \texttt{BER} and then the destination as \texttt{JFK}, but parallel decoding may not see the origin and incorrectly uses \texttt{NYC}. Nevertheless, the impact of such incomplete tool call information on the correctness is minor, because most argument values are taken directly from the user's request and rarely depend on each other. On the BFCL dataset \cite{patil2025bfcl} with Qwen3-1.7B, this impact is as low as 0.9\%, and more detailed results over different LLM families are in Section \ref{section:eval}.

\begin{figure}[t]
	\centering
	\includegraphics[width=\columnwidth]{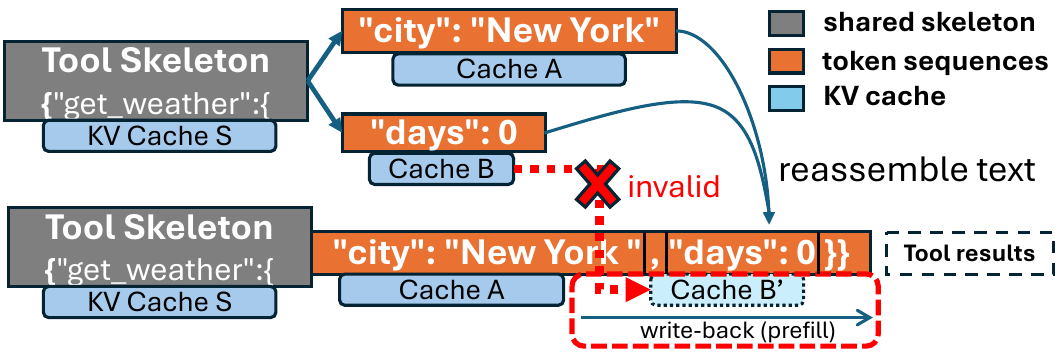}
	\vspace{-0.2in}
	\caption{Parallel decoding invalidates the KV cache entries of token sequences. The write-back reassembles the tool text and rebuilds memory of the whole call.}
	\vspace{-0.25in}
	\label{fig:memory}
\end{figure}

\vspace{-0.05in}
\subsection{Assembling the Tool Call}
\label{sec:assembly}
\label{subsection:call-assembly}

When decoding of all token sequences has finished, the required argument values are all known. However, since different argument values are simultaneously generated in parallel sequences, such parallel decoding invalidates the token sequences' corresponding entries in the LLM agent's KV cache, when these entries are integrated into a single sequence to complete the tool call\footnote{Since KV cache is a sequential data structure \cite{pope2023scaling, kwon2023pagedattention}, altering certain tokens would invalidate all cache entries at later positions in the KV cache.}. For example in Figure \ref{fig:memory}, one token sequence ends with \texttt{"city":"New York"} and the other ends with \texttt{"days":0}, but neither of them completes the tool call.

\myname tackles this issue with the Tool Call Reassembler, which discards KV caches in diverged token sequences, reassembles the decoding results from parallel token sequences into a single sequence, and recomputes the affected KV cache for the complete tool call. The assembled argument texts are then written back to the KV cache in a single prefill pass. %
For tool calls in the BFCL dataset \cite{patil2025bfcl}, the write-back prefills 34.5 tokens on average and takes 522\,ms on the OnePlus 12R, about the time of 9 decode passes.

However, note that this write-back may not increase the end-to-end tool call latency, because the recomputed KV cache is only needed in the next conversation of LLM agent (e.g., when the LLM agent reads the tool's result and writes its reply to the user). Hence, if the tool call has no optional argument, the assembled tool call texts can be handed to the tool for execution before write-back. Calling a tool whose schema contains optional arguments, on the other hand, will be discussed in Section \ref{section:late}.

\section{Handling Optional Arguments}
\label{section:late}
\label{subsection:late-decision}
\label{subsection:provisional-calls}
\begin{figure}[t]
	\centering
	\includegraphics[width=\columnwidth]{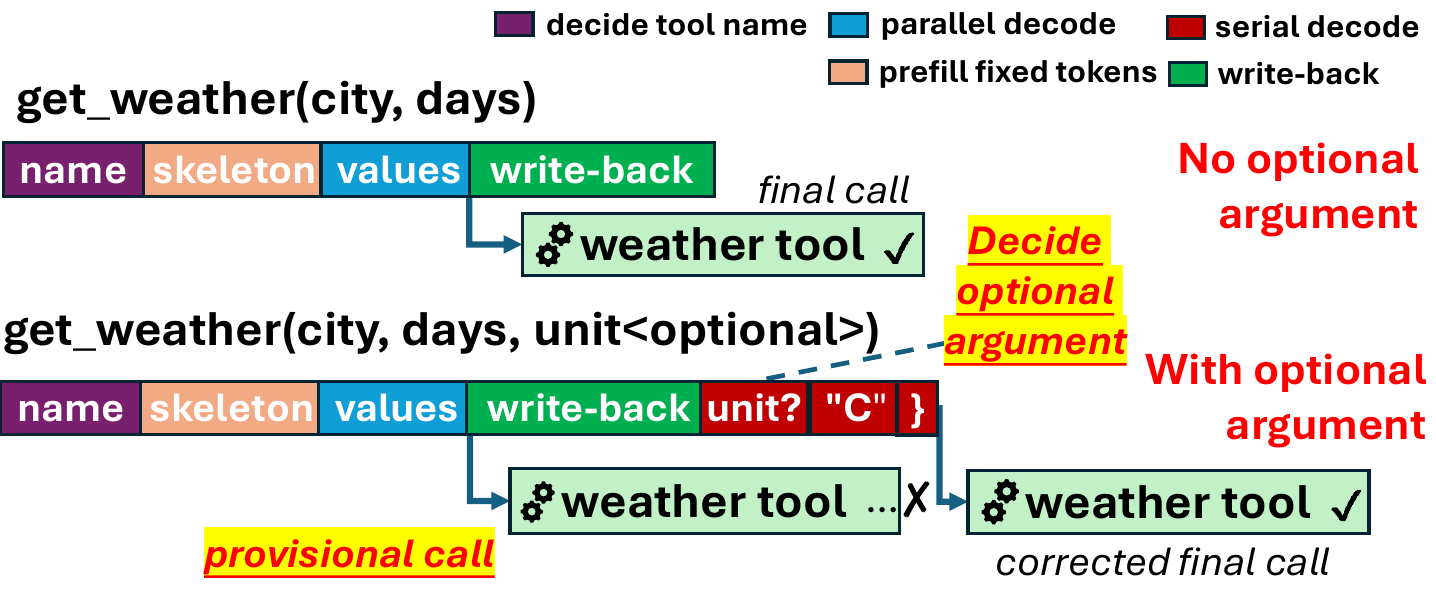}
	\vspace{-0.15in}
	\caption{Handling optional arguments with a provisional call that only contains the required arguments}
	\vspace{-0.1in}
	\label{fig:latecall}
\end{figure}

Many tool calls also contain optional arguments (e.g., 66\% in the BFCL dataset \cite{patil2025bfcl}). %
Figure~\ref{fig:latecall} illustrates the two cases that \myname{} handles, namely tools without and with optional arguments. We first explain when the optional arguments should be decided, and then how provisional calls are designed to minimize the tool call latency with optional arguments.

\vspace{0.05in}
\noindent\textbf{Deciding optional arguments.} For a specific tool call, while its required arguments are pre-defined by tool schema and can be decoded in parallel based on such information, the optional arguments being involved depend on values of required arguments and cannot be decided in advance.

Table~\ref{tab:optional} shows why the time of deciding the involved optional arguments matters. If the optional arguments are decided before any value of required arguments is generated, the correctness of deciding the optional arguments is 17\% lower than that with serial decoding, as the LLM agent has to make guesses in many cases. The correctness of the tool call, correspondingly, also drops by 13.5\%. Another choice is to include all the available optional arguments into the tool skeleton but usually involves unused arguments, further reducing the argument correctness by another 5.2\%.

Hence, \myname opts to decide the optional arguments after values of required arguments are decoded, using serial decoding, and
such late decision of optional arguments improves the argument correctness to 79\%. On the BFCL dataset \cite{patil2025bfcl}, such serial decoding adds 13.5 decode passes on average.

\begin{table}[ht]
	\centering
	\footnotesize
	\setlength{\tabcolsep}{3pt}
	\vspace{-0.1in}
	\begin{tabular}{l|c|c|c}
		\toprule
		\makecell[l]{\textbf{When the optional}\\\textbf{arguments are decided}} & \makecell{\textbf{Argument}\\\textbf{correctness}} & \makecell{\textbf{Tool call}\\\textbf{correctness}} & \makecell{\textbf{Latency}\\\textbf{(ms)}} \\
		\toprule
		End of the call (serial decoding) & 80\% & 76.0\% & 2,586 \\
		\midrule
		\makecell[l]{Before required argument \\ values (in the skeleton)} & 63\% & 62.5\% & 1,105 \\
		\midrule
		Never (all put into the skeleton) & 57\% & 70.8\% & 1,103 \\
		\midrule
		\makecell[l]{After required argument \\ values (\myname)} & 79\% & 74.0\% & 1,211 \\
		\bottomrule
	\end{tabular}
	\vspace{0.02in}
	\caption{Different choices of times when the optional arguments are decided, with the BFCL dataset \cite{patil2025bfcl} and Qwen3-1.7B model}
	\vspace{-0.25in}
	\label{tab:optional}
\end{table}

\noindent\textbf{Provisional calls.} To tackle such extra latency caused by serial decoding of optional argument values, our basic approach is to let the tool start on a provisional call that carries only the required arguments, and the optional arguments are decided and decoded during this provisional tool's execution. As shown in Figure~\ref{fig:latecall}, the write-back described in Section~\ref{subsection:call-assembly} comes after the provisional call and before the optional argument decision.
In this way, both the write-back and the decision of optional arguments can be conducted while the tool executes, hence avoiding the tool execution being delayed.

Afterwards, if no optional argument is involved, the provisional call is considered as correct and its execution results are used to generate the response to the user. Otherwise, the provisional call's execution results will be replaced by the corrected call with the optional arguments added, and the provisional call and its execution results are discarded.

\vspace{0.05in}
\noindent\textbf{Read-only tools.} Since a provisional call may be discarded, it only applies to tools whose execution can be repeated without effect. For example, the tool annotations of the MCP specification mark such a tool as \emph{read-only}, defined as a tool that ``does not modify its environment'' \cite{mcp2025spec}. \myname sends provisional calls only to these read-only tools. In other cases, a tool without read-only annotations (e.g., non-MCP tools) is classified by its name and description. For example, a tool that modifies the local system's configurations is not read-only and cannot be started with a provisional call.

\begin{table*}[t]
	\centering
	\footnotesize
	\setlength{\tabcolsep}{3pt}
	\vspace{-0.1in}
	\begin{tabular}{|l||cc|cc|cc|cc|cc|cc|cc|}
		\hline
		& \multicolumn{2}{c|}{\textbf{Qwen3-1.7B}} & \multicolumn{2}{c|}{\textbf{Qwen3-1.7B 4-bit}} & \multicolumn{2}{c|}{\textbf{Qwen3-0.6B}} & \multicolumn{2}{c|}{\textbf{Qwen3-4B}} & \multicolumn{2}{c|}{\textbf{Llama-3.2-1B}} & \multicolumn{2}{c|}{\textbf{Llama-3.2-3B}} & \multicolumn{2}{c|}{\textbf{SmolLM2-1.7B}} \\
		\hline\hline
		\textbf{Latency (ms)} & 12R & Pixel 7 & 12R & Pixel 7 & 12R & Pixel 7 & 12R & Pixel 7 & 12R & Pixel 7 & 12R & Pixel 7 & 12R & Pixel 7 \\
		\hline
		No Grammar & 2,602 & 3,565 & 2,242 & 2,482 & 1,271 & 1,569 & 5,605 & 24,004 & 1,854 & 2,707 & 4,137 & 6,145 & 3,749 & 5,664 \\
		Constrained decoding & 2,586 & 3,561 & 2,173 & 2,456 & 1,183 & 1,569 & 5,574 & 23,871 & 1,636 & 2,415 & 4,009 & 5,852 & 3,541 & 5,619 \\
		Jump-forward decoding & 1,921 & 2,723 & 1,807 & 1,794 & 1,071 & 1,346 & 3,867 & 18,563 & 1,304 & 1,842 & 2,719 & 4,365 & 2,396 & 3,536 \\
		Speculative, EAGLE-3 & 9,610 & 12,031 & 8,141 & 9,102 & 3,709 & 4,812 & 14,714 & OOM & 8,703 & 12,847 & 14,585 & 21,290 & / & / \\
		Speculative, DFlash-2 & / & / & / & / & / & / & 13,433 & OOM & / & / & / & / & / & / \\
		\textbf{\myname{}} & \textbf{1,211} & \textbf{1,600} & \textbf{1,236} & \textbf{1,272} & \textbf{577} & \textbf{705} & \textbf{2,685} & \textbf{10,334} & \textbf{773} & \textbf{1,101} & \textbf{1,840} & \textbf{2,707} & \textbf{1,498} & \textbf{2,243} \\
		\hline\hline
		\textbf{\# of Forward passes} & \multicolumn{2}{c|}{} & \multicolumn{2}{c|}{} & \multicolumn{2}{c|}{} & \multicolumn{2}{c|}{} & \multicolumn{2}{c|}{} & \multicolumn{2}{c|}{} & \multicolumn{2}{c|}{} \\
		\hline
		No Grammar & \multicolumn{2}{c|}{44.2} & \multicolumn{2}{c|}{43.4} & \multicolumn{2}{c|}{42.8} & \multicolumn{2}{c|}{45.0} & \multicolumn{2}{c|}{45.5} & \multicolumn{2}{c|}{40.8} & \multicolumn{2}{c|}{60.1} \\
		Constrained decoding & \multicolumn{2}{c|}{44.3} & \multicolumn{2}{c|}{41.6} & \multicolumn{2}{c|}{42.4} & \multicolumn{2}{c|}{44.7} & \multicolumn{2}{c|}{42.9} & \multicolumn{2}{c|}{40.0} & \multicolumn{2}{c|}{57.8} \\
		Jump-forward decoding & \multicolumn{2}{c|}{21.9} & \multicolumn{2}{c|}{19.2} & \multicolumn{2}{c|}{19.7} & \multicolumn{2}{c|}{22.4} & \multicolumn{2}{c|}{20.1} & \multicolumn{2}{c|}{18.2} & \multicolumn{2}{c|}{21.0} \\
		Speculative, EAGLE-3 & \multicolumn{2}{c|}{18.1} & \multicolumn{2}{c|}{17.2} & \multicolumn{2}{c|}{13.9} & \multicolumn{2}{c|}{16.9} & \multicolumn{2}{c|}{17.4} & \multicolumn{2}{c|}{16.9} & \multicolumn{2}{c|}{/} \\
		Speculative, DFlash-2 & \multicolumn{2}{c|}{/} & \multicolumn{2}{c|}{/} & \multicolumn{2}{c|}{/} & \multicolumn{2}{c|}{9.2} & \multicolumn{2}{c|}{/} & \multicolumn{2}{c|}{/} & \multicolumn{2}{c|}{/} \\
		\textbf{\myname{}} & \multicolumn{2}{c|}{\textbf{13.1}} & \multicolumn{2}{c|}{\textbf{13.2}} & \multicolumn{2}{c|}{\textbf{12.3}} & \multicolumn{2}{c|}{\textbf{12.7}} & \multicolumn{2}{c|}{\textbf{12.6}} & \multicolumn{2}{c|}{\textbf{11.6}} & \multicolumn{2}{c|}{\textbf{14.5}} \\
		\hline
		\textbf{Correctness} & \multicolumn{2}{c|}{} & \multicolumn{2}{c|}{} & \multicolumn{2}{c|}{} & \multicolumn{2}{c|}{} & \multicolumn{2}{c|}{} & \multicolumn{2}{c|}{} & \multicolumn{2}{c|}{} \\
		\hline
		No Grammar & \multicolumn{2}{c|}{76.0\%} & \multicolumn{2}{c|}{74.2\%} & \multicolumn{2}{c|}{62.9\%} & \multicolumn{2}{c|}{78.1\%} & \multicolumn{2}{c|}{8.0\%} & \multicolumn{2}{c|}{29.0\%} & \multicolumn{2}{c|}{20.0\%} \\
		Constrained decoding & \multicolumn{2}{c|}{76.0\%} & \multicolumn{2}{c|}{73.2\%} & \multicolumn{2}{c|}{65.2\%} & \multicolumn{2}{c|}{78.1\%} & \multicolumn{2}{c|}{60.6\%} & \multicolumn{2}{c|}{78.0\%} & \multicolumn{2}{c|}{65.3\%} \\
		Jump-forward decoding & \multicolumn{2}{c|}{76.0\%} & \multicolumn{2}{c|}{73.2\%} & \multicolumn{2}{c|}{65.2\%} & \multicolumn{2}{c|}{79.2\%} & \multicolumn{2}{c|}{61.6\%} & \multicolumn{2}{c|}{78.0\%} & \multicolumn{2}{c|}{65.3\%} \\
		Speculative, EAGLE-3 & \multicolumn{2}{c|}{76.0\%} & \multicolumn{2}{c|}{73.2\%} & \multicolumn{2}{c|}{65.2\%} & \multicolumn{2}{c|}{78.1\%} & \multicolumn{2}{c|}{60.6\%} & \multicolumn{2}{c|}{78.0\%} & \multicolumn{2}{c|}{/} \\
		Speculative, DFlash-2 & \multicolumn{2}{c|}{/} & \multicolumn{2}{c|}{/} & \multicolumn{2}{c|}{/} & \multicolumn{2}{c|}{78.1\%} & \multicolumn{2}{c|}{/} & \multicolumn{2}{c|}{/} & \multicolumn{2}{c|}{/} \\
		\textbf{\myname{}} & \multicolumn{2}{c|}{74.0\%} & \multicolumn{2}{c|}{72.1\%} & \multicolumn{2}{c|}{63.6\%} & \multicolumn{2}{c|}{81.2\%} & \multicolumn{2}{c|}{58.0\%} & \multicolumn{2}{c|}{76.0\%} & \multicolumn{2}{c|}{64.0\%} \\
		\hline
	\end{tabular}
	\par\smallskip\noindent{\scriptsize /: no draft model or head exists for the model. OOM: out of memory on the phone.}
	\vspace{0.05in}
	\caption{The end-to-end latency and correctness of tool calls over different LLMs. BFCL-v4 dataset is used.}
	\vspace{-0.2in}
	\label{tab:latency}
\end{table*}

\begin{figure}[ht]
	\centering
	\includegraphics[width=\columnwidth]{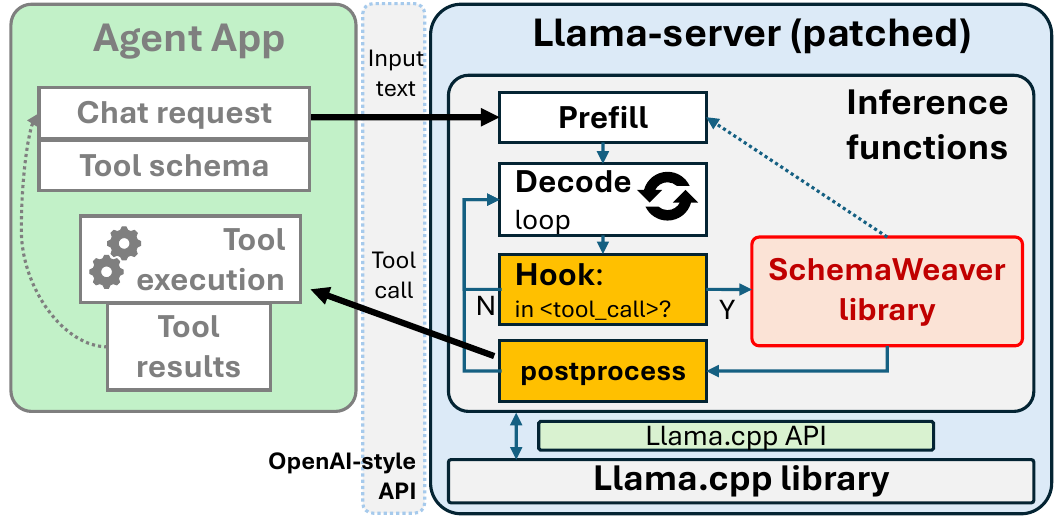}
	\vspace{-0.25in}
	\caption{Implementation of \myname{} inside the llama-server}
	\vspace{-0.1in}
	\label{fig:impl}
\end{figure}

\vspace{-0.05in}
\section{Implementation}
\label{section:setup}

We implemented \myname{} on llama.cpp~\cite{llamacpp} (commit \texttt{d59d455f}, August 2026), the most commonly used open-source LLM inference engine for Android smartphones. Our implementation consists of a library of 930 lines of C++ code on the public API of llama.cpp, together with a 350-line patch to its server as a hook on tool calls to its decode loop. When the LLM agent chooses to emit a tool call, the hook lets \myname{} produce the call, and returns the provisional call to the agent app as soon as it is assembled, before the optional arguments are decided. When the call is complete, the hook feeds its tokens back to the server exactly as if they had been decoded in serial. The engine itself is otherwise unchanged. Figure~\ref{fig:impl} shows how the pieces fit together.

\myname{} is deployed on two smartphone models with different mobile SoCs, i.e., a OnePlus 12R with Qualcomm Snapdragon 8 Gen 2 SoC and 16\,GB of memory running Android 14, and a Google Pixel 7 with Google Tensor G2 SoC and 8\,GB of memory running Android 16. The LLM inference engine and \myname{} are built from source with \texttt{clang} 21.1 in Termux, and we use the publicly released GGUF files of LLMs in 8-bit (Q8\_0) and 4-bit (Q4\_0) quantization. Both smartphones are controlled from a desktop over the Android Debug Bridge (adb), which launches \myname{} and the baselines to collect experiment data.

\vspace{-0.05in}
\section{Evaluation}
\label{section:eval}

In our evaluation, we use open-sourced LLMs including Qwen3~\cite{yang2025qwen3}, Llama-3.2~\cite{grattafiori2024llama3} and SmolLM2~\cite{allal2025smollm2}. The model size varies from 0.6B to 4B to fit the memory and compute limits of smartphones. We use the following tool call datasets to evaluate the on-device performance of tool calling:
\begin{itemize}
	\item \textbf{BFCL-v4}~\cite{patil2025bfcl} consists of 2,501 single-turn tool call samples across 991 tools, with up to 9 arguments per call and 2.4 on average. Our evaluation uses a subset of 100 calls, uniformly sampled over the number of arguments per call.
	\item \textbf{DroidCall}~\cite{xie2024droidcall} is a dataset for Android system intents, i.e., tool calls that open apps, set alarms or send messages on the phone, from which we take its 154 single-call requests over 23 tools.
\end{itemize}

\noindent\textbf{Experiment protocol.} To ensure fair comparisons, the inference engine process is pinned to 4 performance cores of mobile CPU, at a fixed CPU clock under a cooling fan, since thermal throttling would otherwise change CPU frequency in the middle of a run. The choice of the core set and CPU clock is given in Appendix~\ref{app:cores}. %
Peak resident memory is read from the mobile processor's VmHWM counter polled once per second. Time consumption is measured via the phone's builtin monotonic clock.

\noindent\textbf{Baselines.} We compare the performance of \myname{} with the following four decoding schemes, and every baseline runs in the same engine software build as \myname.
\begin{itemize}
	\item \textbf{Unconstrained decoding}, shown as \emph{No Grammar} in tables and figures, lets LLM write the call freely.
	\item \textbf{Constrained decoding}~\cite{willard2023outlines, geng2023gcd, dong2024xgrammar, llamacpp} applies the schema constraint built into llama.cpp on top of unconstrained decoding.
	\item \textbf{Jump-forward decoding}~\cite{zheng2024sglang} skips the tokens that the schema fixes when the decoder meets them, as described in Section~\ref{subsection:call-skeleton}.
	\item \textbf{Speculative decoding}~\cite{leviathan2023speculative} guesses the next tokens using a smaller draft model for the target model to check its correctness. Due to the limitation in available draft and target models in LLM families we use, we test two different schemes with different models: 1) EAGLE-3~\cite{li2025eagle3} with a draft layer of 59M to 247M parameters that reads the target model's hidden states, available for every model in our set except SmolLM2; 2) DFlash-2~\cite{chen2026dflash} with Qwen3-4B as target model and a 1.0\,GB block-diffusion draft model.
\end{itemize}

\noindent\textbf{Metrics.} Our primary evaluation metric is the \emph{tool call latency}. For \myname, we compute the latency as the elapsed time between the start of generating tool call texts and the start of tool execution.
For the baselines, the latency ends at the last token of the tool call. Latency is reported as the geometric mean~\cite{fleming1986geomean, hennessy2019caqa} over the repetitions of each call and then over the calls, since it is the mean that summarizes ratios consistently. We report a speedup as the ratio of two such means.

Besides, we also evaluate the whole-call \emph{correctness}, i.e., the percentage of the tool calls made whose tool and arguments match the accepted answers in the dataset. The number of forward passes per call, the peak resident memory and energy cost per call are also evaluated and reported as more detailed information about the compute costs of tool calls.

\subsection{Latency and Correctness of Tool Calling}
\label{sec:eval-latency}

We first compare \myname{} with the four baselines on the BFCL dataset. As shown in Table~\ref{tab:latency}, across the seven model settings and on both phones, \myname{} keeps the correctness of tool calls at the same level as constrained decoding: on Qwen3-1.7B, the whole-call correctness is 76.0\% under constrained decoding and 74.0\% with \myname{}. We find that \myname{} chooses the same arguments as constrained decoding does, and the correctness loss comes from parallel decoding, where the values cannot see each other. Jump-forward decoding produces the same tokens as constrained decoding and keeps its correctness. The same applies to the speculative rows: on the Qwen3 models, decoding with No Grammar stays within about 2\% difference from that of constrained decoding, so the constraint costs no correctness. The Llama and SmolLM2 models need the grammar for the call format itself. Without it they easily write the call in the wrong format, and their correctness falls to 8--29\%.

Given such correctness of tool calling, \myname{} significantly reduces the end-to-end latency of tool calling. On Qwen3-1.7B, its end-to-end latency is 1,211\,ms on the OnePlus 12R smartphone, a $2.13\times$ speedup over constrained decoding. The number of forward passes is correspondingly reduced from 44.3 to 13.1.  Jump-forward decoding reaches $1.82\times$, since it skips the fixed tokens but still decodes every argument value one token per pass. Decoding with no grammar is as slow as constrained decoding, so the grammar constraint is not the bottleneck. \myname{} reaches a higher speedup of $2.36\times$ on SmolLM2-1.7B, since its call format consists of more fixed tokens.

The speculative rows gain little speedup on both phone models and all LLM families. %
EAGLE-3 for Qwen3-1.7B and DFlash-2 for Qwen3-4B run slower than constrained decoding at $0.27\times$ and $0.42\times$, respectively. Two costs on the phone turn the gain into a loss. The verification pass in speculative decoding checks the drafted tokens as one batched pass, which on the phone costs more than a serial step according to Table~\ref{tab:batchcost}. With 4.5 of the 16 drafted tokens accepted per block, the verification pass of DFlash-2 alone costs more than the serial steps it replaces. The draft head also reads its own weights from the same DRAM at every drafted token, so the drafting competes for the bandwidth that limits the decode phase. Besides, the EAGLE-3 head must process the whole prompt during prefill, from 11.4 to 50.8\,ms per prompt token, and that accounts for three quarters of its loss.

\begin{table}[t]
	\centering
	\footnotesize
	\setlength{\tabcolsep}{3pt}
	\begin{tabular}{|l||cc|cc|cc|}
		\hline
		& \multicolumn{2}{c|}{\textbf{Qwen3-1.7B}} & \multicolumn{2}{c|}{\textbf{Llama-3.2-1B}} & \multicolumn{2}{c|}{\textbf{SmolLM2-1.7B}} \\
		\hline\hline
		\textbf{Latency (ms)} & 12R & Pixel 7 & 12R & Pixel 7 & 12R & Pixel 7 \\
		\hline
		Constrained & 1,771 & 2,523 & 1,035 & 1,567 & 2,682 & 4,136 \\
		Jump-forward & 804 & 1,080 & 461 & 693 & 1,115 & 1,630 \\
		\textbf{\myname{}} & \textbf{677} & \textbf{944} & \textbf{328} & \textbf{490} & \textbf{779} & \textbf{1,166} \\
		\hline\hline
		\textbf{Correctness} & \multicolumn{2}{c|}{} & \multicolumn{2}{c|}{} & \multicolumn{2}{c|}{} \\
		\hline
		Constrained & \multicolumn{2}{c|}{67.2\%} & \multicolumn{2}{c|}{58.4\%} & \multicolumn{2}{c|}{69.3\%} \\
		Jump-forward & \multicolumn{2}{c|}{67.2\%} & \multicolumn{2}{c|}{58.4\%} & \multicolumn{2}{c|}{68.6\%} \\
		\textbf{\myname{}} & \multicolumn{2}{c|}{67.2\%} & \multicolumn{2}{c|}{61.7\%} & \multicolumn{2}{c|}{67.4\%} \\
		\hline
	\end{tabular}
	\vspace{0.05in}
	\caption{The end-to-end latency and correctness of tool calls over different LLMs. DroidCall dataset is used.}
	\vspace{-0.3in}
	\label{tab:droidcall}
\end{table}

Table~\ref{tab:droidcall} further shows the experiment results on the DroidCall dataset, whose tools are Android system functions with 1.5 arguments per call on average. \myname{} reduces the tool call latency from 1,771\,ms of constrained decoding to 677\,ms on the OnePlus 12R with Qwen3-1.7B, reaching a speedup of $2.62\times$, and holds between $2.67\times$ and $3.55\times$ in the other settings, with correctness loss against constrained decoding controlled within 2\%. The gain is larger than that on the BFCL dataset although the calls are shorter, because most of these calls have one required argument and five of the 23 tools are read-only, and so the provisional call handles the tool early as soon as the required value is decoded.

\begin{figure}[t]
	\centering
	\includegraphics[width=0.95\columnwidth]{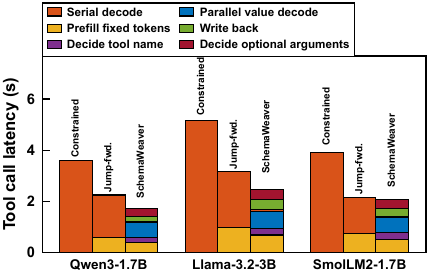}
	\vspace{-0.1in}
	\caption{Tool call latency by phase on the OnePlus 12R}
	\vspace{-0.2in}
	\label{fig:phases}
\end{figure}

\subsection{Different Model Sizes and Quantizations}
\label{sec:eval-models}

Table~\ref{tab:latency} lists the measured latencies across six models, one of them (Qwen3-1.7B) in two quantizations. The speedup over constrained decoding holds between $1.76\times$ and $2.36\times$ on the OnePlus 12R and between $1.93\times$ and $2.50\times$ on the Pixel 7. To see where the time of a call goes under each setup, Figure~\ref{fig:phases} splits the latency of the three schemes by phase on one model of each family. On Qwen3-1.7B and calls with 2+ required values, the tool name takes 170\,ms, the skeleton pass 351\,ms and the parallel values 572\,ms on average, while the write-back (522\,ms) and the optional arguments (247\,ms) fall after the provisional call and overlap the tool execution for read-only tools. Llama~3.2 and SmolLM2 spend more fixed tokens per call due to their call formats, so their skeletons replace more serial decoding and their speedups in Table~\ref{tab:latency} are larger.

The skeleton prefill costs about the same (5 to 6 serial decode steps) in every setting, so the differences in speedup come from the call format and the quantization. The 4-bit Qwen3-1.7B gains the least. Its weights are half the size, but its decode step is only 13\% faster, because unpacking 4-bit weights costs extra arithmetic. Qwen3-4B does not fit in the Pixel 7's 8\,GB of memory. The engine keeps up to 1.5\,GB of the 4.3\,GB model in compressed swap in RAM, so a call under constrained decoding takes significantly longer (23.9\,s) due to swapping. Even so, \myname{} still achieves a speedup of $2.31\times$.

On the full BFCL dataset, \myname{} loses 0.7 to 2.0 points of whole-call correctness across the seven settings, with the largest loss on Llama-3.2-1B. The loss comes from the required values decoded in parallel, which cannot see each other. On the 100-call subset in Table~\ref{tab:latency}, the differences vary more ($+3.1$ to $-5.6$ points) since the subset is small, but they remain within a few points.

\begin{figure}[t]
	\centering
	\includegraphics[width=\columnwidth]{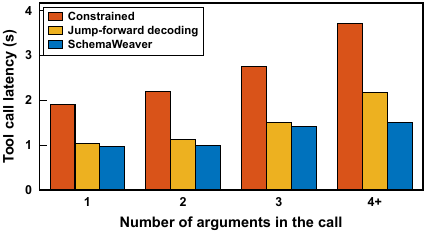}
	\vspace{-0.2in}
	\caption{Tool call latency by the number of arguments of the call on the OnePlus 12R.}
	\vspace{-0.2in}
	\label{fig:args}
\end{figure}

\subsection{Impact of Memory Bandwidth}
\label{sec:eval-bandwidth}

\begin{figure}[t]
	\centering
	\begin{subfigure}[t]{0.49\columnwidth}
		\centering
		\includegraphics[width=\linewidth]{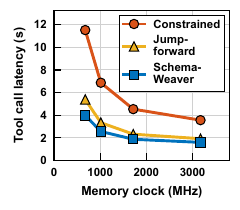}
		\caption{Latency}
	\end{subfigure}
	\begin{subfigure}[t]{0.49\columnwidth}
		\centering
		\includegraphics[width=\linewidth]{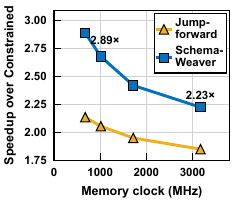}
		\caption{Speedup}
	\end{subfigure}
	\vspace{-0.1in}
	\caption{Tool call latency and speedup with different memory bus frequencies on the Pixel 7.}
	\vspace{-0.1in}
	\label{fig:bandwidth}
\end{figure}

Section~\ref{subsec:bandwidth} shows that the decode phase is bound by memory bandwidth, so the gain of \myname{} should grow on phones with a slower memory bus. To verify this, we pin the memory bus of the Pixel 7 at four of its own frequency levels, from the stock maximum of 3,172\,MHz down to 676\,MHz, with the CPU clock pinned and the fan on. We repeat the BFCL calls of Table~\ref{tab:latency} with Qwen3-1.7B at every level. Figure~\ref{fig:bandwidth} plots the latency of every scheme and the speedup of \myname{} against the memory clock, i.e., the frequency of the memory bus.

The gain of \myname{} indeed grows as memory bandwidth falls. As the bus slows $4.7\times$, constrained decoding slows $3.23\times$ while \myname{} slows $2.49\times$, so its speedup rises from $2.23\times$ at the stock clock to $2.89\times$ at 676\,MHz. Jump-forward decoding also gains from the slower bus, since the tokens it skips would have cost decode steps as well, but it gains less ($1.85\times$ to $2.14\times$), so the gap between its speedup and that of \myname{} grows from 0.38 to 0.75.

The per-phase records explain this difference. With the lowered bus frequency, a serial decode step slows $3.23\times$, from 90 to 291\,ms, because every step reads the whole model from memory. The four-sequence value pass slows almost as much, from 107 to 332\,ms, since it reads the weights once as well. The prefill passes behave differently: the skeleton prefill slows only $1.62\times$, from 417 to 674\,ms, and a prompt token only $1.12\times$, because a prefill reads the weights once for many tokens and stays compute-bound. \myname{} replaces most of the serial steps of a call with prefill work and a few batched passes, so a smaller part of its latency scales with the bus.

\subsection{Impact of the Number of Arguments}
\label{sec:eval-tools}

The gain of \myname{} grows with the number of arguments of the call, since every argument adds to the decode parallelism. Figure~\ref{fig:args} shows the latency by the number of arguments. A one-argument call has nothing to decode in parallel, and its $1.9\times$ speedup comes from the skipped fixed tokens and the provisional call alone, so the gap to jump-forward decoding is the smallest there. Calls with four or more arguments gain $2.5\times$, since the parallel pass batches the most values there, and the gap to jump-forward decoding is the largest. The three-argument calls gain less than the two-argument ones ($1.9\times$ against $2.2\times$), because the model adds an optional argument after the provisional call in 72\% of them, against about half of the other calls, so their tool call is more often the corrected call that waits for the write-back and the serial decision.

\begin{table}[t]
	\centering
	\footnotesize
	\setlength{\tabcolsep}{3.5pt}
	\begin{tabular}{lrr}
		\toprule
		Setup & Latency (ms) & Correctness \\
		\midrule
		Constrained decoding & 2,586 & 76.0\% \\
		+ skip the fixed tokens (jump-forward) & 1,421 & 76.0\% \\
		\quad jump-forward with a provisional call & 1,349 & 76.0\% \\
		+ skeleton and provisional call & 1,369 & 75.0\% \\
		+ parallel value sequences & 1,247 & 74.0\% \\
		+ padded value batch (= \myname{}) & \textbf{1,211} & 74.0\% \\
		\midrule
		Argument set given (upper bound) & 1,036 & 77.1\% \\
		\bottomrule
	\end{tabular}
	\vspace{0.05in}
	\caption{Latency and correctness on \myname{} design variants}
	\vspace{-0.2in}
	\label{tab:ablation}
\end{table}

\subsection{Ablation Studies}
\label{sec:eval-ablation}

Table~\ref{tab:ablation} adds the design steps of \myname{} one at a time on the OnePlus 12R with Qwen3-1.7B. The first four rows separate the gain of skipping the fixed tokens from the gain of the parallel values. Skipping the fixed tokens, by jump-forward decoding or by the tool skeleton, together with the provisional call reaches a speedup of up to $1.92\times$. Parallel decoding of the argument values and the padded batch then raise the speedup to $2.13\times$. The skeleton and the parallel values cost one point of whole-call correctness each, and the padding costs none. The last row is an upper bound. When the set of arguments is given in advance, the call starts 15\% sooner and is 3.1 points more correct than with \myname{}, so deciding the optional arguments late recovers most of that correctness.

\begin{table}[ht]
	\centering
	\footnotesize
	\begin{tabular}{lrrr}
		\toprule
		Phase & Passes & Tokens & Time (ms) \\
		\midrule
		Constrained decoding & 35 & 35 & 2,004 \\
		\midrule
		Tool selector & 1 & 6 & 109 \\
		Skeleton prefill & 1 & 19 & 268 \\
		Parallel decode & 2 & 3 & 128 \\
		\quad Provisional tool call & & & \textbf{505} \\
		Write-back & 1 & 18 & 238 \\
		Optional argument & 4 & 7 & 238 \\
		\quad Final tool call & & & \textbf{981} \\
		\bottomrule
	\end{tabular}
	\vspace{0.05in}
	\caption{Cost breakdown for \texttt{get\_traffic\_info} tool, on the OnePlus 12R with Qwen3-1.7B.}
	\vspace{-0.1in}
	\label{tab:example}
\end{table}

Table~\ref{tab:example} breaks down the latency of one of the 100 BFCL calls by phase. The user asks for the traffic from Boston to New York, and the tool \texttt{get\_traffic\_info} has 2 required arguments and 1 optional argument. Constrained decoding spends 35 decode passes on the call. \myname{} decides the name in one pass, prefills the skeleton in one pass, decodes the two values in two passes, and emits the provisional call to the tool after 505\,ms, $3.97\times$ sooner than constrained decoding. The write-back and the optional argument, decided as Section \ref{section:late} describes, complete the final call after 981\,ms, with 9 passes instead of 35 and the same call text as constrained decoding produces.

\vspace{-0.05in}
\subsection{Memory Overhead}
\label{sec:eval-memory}

\myname{} is a memory-efficient approach, especially compared with designs based on speculative decoding. Table~\ref{tab:memory} lists the peak memory of each scheme. The peak resident memory for \myname{} on the OnePlus 12R is only 8.9\,MB above constrained decoding on Qwen3-1.7B. This extra memory holds the KV cache entries of the parallel sequences and their type constraints. Decoding with no grammar and jump-forward decoding add less than 1\,MB, since neither keeps any state beyond the serial decoder's. The EAGLE-3 head adds about 2.2\,GB on top of its own constrained-decoding run, for its weights and for the prompt it must process alongside the target model.

\begin{table}[ht]
	\centering
	\scriptsize
	\setlength{\tabcolsep}{2.5pt}
	\begin{tabular}{|l||c|c|c|c|c|}
		\hline
		& \makecell{\textbf{Constrained}\\\textbf{decoding}} & \makecell{\textbf{No}\\\textbf{Grammar}} & \makecell{\textbf{Jump-}\\\textbf{forward}} & \makecell{\textbf{Speculative,}\\\textbf{EAGLE-3}} & \textbf{\myname{}} \\
		\hline\hline
		\makecell[l]{Peak\\memory} & 4.725\,GB & 4.725\,GB & 4.725\,GB & 6.926\,GB & \textbf{4.734\,GB} \\
		\hline
	\end{tabular}
	\vspace{0.05in}
	\caption{Peak resident memory of each scheme on the OnePlus 12R with Qwen3-1.7B.}
	\vspace{-0.2in}
	\label{tab:memory}
\end{table}

\vspace{-0.2in}
\subsection{Energy Consumption}
\label{sec:eval-energy}

\begin{figure}[ht]
	\centering
	\includegraphics[width=0.95\columnwidth]{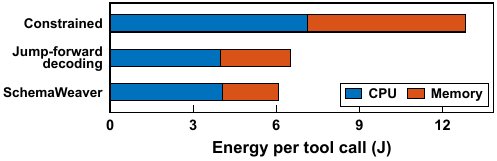}
	\vspace{-0.1in}
	\caption{Energy per tool call on the Google Pixel 7}
	\vspace{-0.2in}
	\label{fig:energy}
\end{figure}

We further evaluate the energy consumption of \myname{} per tool call. Figure~\ref{fig:energy} shows the energy cost of a tool call on the Pixel 7, split into the CPU rails and the memory rails. \myname{} reduces the energy of a tool call from 12.8 to 6.1\,J, i.e., by $2.10\times$. Jump-forward decoding saves almost as much ($1.97\times$), but the savings come from different rails. Jump-forward decoding saves memory energy only on the tokens it skips ($2.3\times$). \myname{} saves $2.8\times$ on the memory rails, since one parallel pass reads the weights once for up to four values, but spends part of that saving on the CPU rails for the write-back, the optional arguments and the padding. The energy of a call counts the write-back and the optional arguments even when they overlap the tool execution. For nearly the same energy, \myname{} starts the tool call $2.23\times$ sooner than constrained decoding, against $1.85\times$ for jump-forward decoding.

\section{Related Work}
\label{section:related}

\noindent\textbf{On-device LLM agents.} Many mobile agents operate existing applications through the GUI~\cite{zhang2023appagent, wen2024autodroid, wen2024autodroidv2, wang2024mobileagent}. Mobile platforms are recently adding native, schema-described tools, such as Android AppFunctions~\cite{android2026appfunctions, android2026intelligentos}, iOS App Intents~\cite{apple2024appintents} and MCP~\cite{hou2025model}. Agent-X~\cite{chung2026agentx} accelerates the on-device agent pipeline by reusing the unchanged parts of the prompt across turns, while \myname{} shortens the generation of the tool call within each turn, so the two can be used together. SPORK and PASTE~\cite{spork2026, paste2026} guess the tool call while the model is still generating it, start the tool early, and discard the result if the guess was wrong, which partially inspires the optional argument handling in \myname{}.

\noindent\textbf{Parallel text generation.} Another line of work generates several segments of text in parallel: Skeleton-of-Thought~\cite{ning2024sot}, APAR and PASTA~\cite{liu2024apar, jin2025pasta} target free-form texts only. SimpleTool focuses on tool calls and decodes a function's name and arguments in parallel from the same independence observation~\cite{shi2026simpletool}. However, it trains special tokens into the model and evaluates on a desktop GPU. HyFunc also shortens function calls, with a trained cascade that hands the call from a large model to a small one and fills the fixed syntax from a template~\cite{liao2026hyfunc}. \myname{} is training-free, runs off-the-shelf quantized models on phones, and grounds its design in their memory-bandwidth profile.

\noindent\textbf{Speculative decoding.} Speculative decoding drafts several tokens with a cheaper mechanism and verifies them in one forward pass of the target model; the draft comes from a smaller model, from small heads trained onto the target model, or from the model's own recent output~\cite{leviathan2023speculative, cai2024medusa, li2024eagle, li2025eagle3, fu2024lookahead}. Newer drafts come from the workload itself: SuffixDecoding and AgentSpec reuse the repetitive outputs of agent pipelines on a serving cluster~\cite{oliaro2025suffixdecoding, wang2026agentspec}, and ToolSpec drafts a tool call from its schema and from retrieved past calls of the same tools~\cite{xia2026toolspec}. These gains rely on a history of similar calls, which the first tool call on a phone does not have, and a wrong draft still costs the batched pass that rejects it. On phones, speculative systems keep the draft model in memory and load the target from flash only to verify a batch of drafted tokens as in LLMCad, EdgeLLM and Lever~\cite{xu2023llmcad, xu2025edgellm, wang2026lever}, or schedule the draft and the verification on different processors~\cite{ma2026ahasd, chen2025npuspec}. Self-speculative decoding drafts with the target model itself by skipping some of its layers, as in Draft \& Verify~\cite{zhang2024draftverify} and SWIFT~\cite{xia2025swift}, so it needs no second model. However, they are designed for larger models (e.g., 13B) that cannot fit onto a phone. In every case, the draft model is a second set of weights that the memory bus must also move. \myname{} needs no second model, and the passes it saves are fixed by the schema rather than by how well a draft guesses. The two can be combined: a draft model shortens each value sequence while \myname{} decodes the sequences together.

\vspace{-0.05in}
\section{Discussions}
\label{section:discussion}

\noindent\textbf{Mobile accelerators.} \myname{} runs on the mobile CPU. The schedule change itself is not CPU-specific, since every pass it adds or removes is an ordinary batched forward pass, but it brings no gain on the phone's other processors today. On the GPU, a pass that advances four value sequences costs 2.5 serial decode steps against 1.2 on the CPU, a serial step is 1.7$\times$ slower than the CPU's at the same clock, and running the whole design there makes the call slower than constrained decoding (0.39$\times$). The contention with the GUI and the longer initialization time on GPU are also problematic. On the NPU, a four-sequence pass costs about three serial steps, and a serial step takes about 3$\times$ longer than on the CPU, since the inference engine keeps the model's output layer on the CPU for now and every decode step then pays a round trip between the two processors. What the accelerators do offer is a cheap prefilled token, 10$\times$ cheaper than a decode step on the GPU and over 70$\times$ on the NPU~\cite{xu2025mllmnpu, chen2025llm}. This matches the skeleton and write-back prefills of \myname{}; letting the NPU run those prefills while the CPU decodes the value sequences is our future work, if moving the state between the two processors costs less than it saves. MNN-AECS~\cite{huang2025mnnaecs} runs the bandwidth-bound decode on fewer and slower CPU cores to save energy; \myname{} shortens that same phase, so the two designs can be combined.

\noindent\textbf{Other inference engines.} \myname{} needs two things from an engine: a schema-derived grammar and attention masks supplied by the caller in the batched decode pass. Only llama.cpp offers both today, while MNN-LLM~\cite{wang2024mnnllm}, ExecuTorch, Google AI Edge SDK~\cite{google2026aiedgefc} and MLC-LLM~\cite{mlcllm} with TVM~\cite{chen2018tvm} miss necessary features on Android. Since the technique behind \myname{} is not engine-specific, we expect similar speedup on other inference frameworks.

\vspace{-0.1in}
\section{Conclusion}
\label{section:conclusion}

In this paper, we present \myname{}, a new technique to accelerate the use of LLM tools on mobile devices by allowing the parallel generation of argument values for tool calls. \myname{} achieves up to $3.9\times$ speedup in tool call latency without noticeable correctness loss compared to the existing schemes, and also reduces energy consumption by 2.1$\times$.

\bibliographystyle{ACM-Reference-Format}
\bibliography{schemaweaver-arxiv}

\newpage
\appendix

\section{The Core Set of the Measurement Protocol}
\label{app:cores}

The measurement protocol of Section~\ref{section:eval} pins one engine thread to each of four CPU cores and fixes the clock of these cores. On the OnePlus 12R, the four cores are the A715 cores at 1.79\,GHz; on the Pixel 7, they are the two X1 cores at 1.745\,GHz and the two A78 cores at 1.663\,GHz. These are the highest clocks that the two phones sustain under continuous load with the cooling fan on. The pinning is necessary. Without it, the Android scheduler stacks two or three of the four threads on one core and leaves another core idle, and the same engine decodes $1.4\times$ to $2.3\times$ slower on the 12R.

Table~\ref{tab:cores} explains the choice of these four cores. It lists the cost of a serial decode step, a four-sequence step and a prefill token on different core sets of the two SoCs, measured with llama.cpp's batched benchmark on Qwen3-1.7B at Q8\_0 with the clocks pinned as above.

\begin{table}[H]
	\centering
	\footnotesize
	\setlength{\tabcolsep}{3pt}
	\begin{tabular}{@{}llrrrrr@{}}
		\toprule
		& & & \multicolumn{2}{c}{Step (ms)} & & Prefill \\
		\cmidrule(lr){4-5}
		Phone & Core set & Threads & serial & 4-sequence & Ratio & ms/token \\
		\midrule
		12R & 2 A715 & 2 & 65.0 & 112.3 & 1.73 & 20.1 \\
		12R & \textbf{4 A715} & 4 & 54.2 & 65.9 & \textbf{1.22} & 9.8 \\
		12R & 4 A715 + X3 & 5 & 43.2 & 50.8 & 1.18 & 5.6 \\
		12R & all 8 & 8 & 59.1 & 73.7 & 1.25 & 6.9 \\
		12R & 3 A510 + X3 & 4 & 73.7 & 98.6 & 1.34 & 13.9 \\
		\midrule
		Pixel 7 & 2 X1 & 2 & 77.4 & 93.2 & 1.21 & 15.9 \\
		Pixel 7 & \textbf{2 X1 + 2 A78} & 4 & 79.2 & 88.0 & \textbf{1.11} & 10.2 \\
		Pixel 7 & all 8 & 8 & 100.9 & 119.9 & 1.19 & 11.4 \\
		Pixel 7 & 4 A55 & 4 & 174.0 & 286.8 & 1.65 & 42.0 \\
		\bottomrule
	\end{tabular}
	\caption{Cost of a serial decode step, a four-sequence step and a prefill token on different core sets of the two phones; Ratio is the four-sequence step over the serial step; the protocol's core set in bold.}
	\label{tab:cores}
\end{table}

Three observations follow. First, the extra cost of a four-sequence step over a serial step is the lowest on the four chosen cores, $1.22\times$ on the 12R and $1.11\times$ on the Pixel~7, and adding more cores does not lower it ($1.18$--$1.25\times$ on the 12R, $1.19\times$ on the Pixel~7). Second, the little cores slow down every step, serial and batched alike. The engine waits for all threads after every layer, so a pass runs at the speed of the slowest core. Using all eight cores is 9\% slower than using four on the 12R and 27\% slower on the Pixel~7, and the four little cores of the Pixel~7 alone are two to three times slower. Third, the largest core of the 12R, the X3, shortens every pass when it is added, the serial step by 20\% and the prefill token by 43\%. However, its clock is not sustainable under continuous load, and it shortens the baseline and \myname{} alike, so the protocol leaves it out. The two X1 cores of the Pixel~7 already saturate its memory bus, since two threads decode as fast as four.

\end{document}